%% file: main.tex
\pdfoutput=1
\documentclass{article}

\usepackage[final]{corl_2026}

\usepackage{makecell}
\input{preamble}

\title{SPARK: Simple Post-training for Adapting pRetrained Knowledge to Robot Control}

\author{
  \aname{Taekyung Kim$^{1}$}%
  \aname{Jeongeun Park$^{2}$}%
  \aname{Byeongho Heo$^{1}$} \\
  \aname{Sungjoon Choi$^{3}$}%
  \aname{Sangdoo Yun$^{1}$}%
  \aname{Dongyoon Han$^{1}$} \\[10pt]
  $^{1}$NAVER AI Lab \quad $^{2}$Georgia Institute of Technology \quad $^{3}$Korea University \\[5pt]
  \texttt{\{taekyung.k, bh.heo, sangdoo.yun, dongyoon.han\}@navercorp.com}\\
  \texttt{jpark3456@gatech.edu} \quad\quad \texttt{sungjoon-choi@korea.ac.kr}
}

\begin{document}
\maketitle
\input{sec/0_abstract}
\input{sec/1_intro}

\input{sec/2_related_work}

\input{sec/3_method}

\input{sec/4_experiments}

\input{sec/5_analysis}
\input{sec/6_conclusion}

\clearpage
\input{sec/x_appendix}

\bibliography{main}

\end{document}

%% file: preamble.tex
\usepackage{multirow}
\usepackage{adjustbox}
\usepackage{amsmath}

\usepackage[utf8]{inputenc}
\usepackage[T1]{fontenc}
\usepackage{url}
\usepackage{booktabs}
\usepackage{amsfonts}
\usepackage{nicefrac}
\usepackage{microtype}
\usepackage{xcolor}
\usepackage{subcaption}
\usepackage{wrapfig}
\PassOptionsToPackage{table}{xcolor}

\definecolor{darkergreen}{RGB}{21, 152, 56}

\newcommand{\aname}[1]{\makebox[0.28\textwidth][c]{\textbf{#1}}}

%% file: sec/0_abstract.tex
\begin{abstract}
Large-scale representation learning has produced public foundation models that provide strong general-purpose visual features.
However, their pretraining objectives are not designed for robot representation learning, so the resulting embeddings tend to emphasize global semantics rather than the temporally sensitive representations required for robot control.
Training robot models from scratch is also undesirable due to the limited scale and high collection cost of robotics data, as well as the substantial computational burden of pre-training.
In this paper, we propose \textbf{SPARK} (\textbf{S}imple \textbf{P}ost-training for \textbf{A}dapting p\textbf{R}etrained \textbf{K}nowledge), a post-training method that adapts foundation models to robot control.
SPARK is built on two principles: temporal abstraction, which encourages the encoder to derive compact features that preserve information essential for understanding temporal changes and action-relevant scene structure, and knowledge preservation, which aligns patch-level representations with those of the original foundation model to retain useful pretrained semantics.
These objectives yield compact visual state representations that remain semantically meaningful while preserving useful prior knowledge.
Experiments across multiple robotics benchmarks show that SPARK consistently improves success rates and generalization over vanilla foundation models, and further demonstrate that these gains transfer to real-world robot manipulation.

\end{abstract}

\keywords{Robot Representation Learning, Post-training}

%% file: sec/1_intro.tex
\section{Introduction}

Robot representation learning~\citep{nair2022r3m, ma2022vip, radosavovic2023mvp, cortexbench, parisi22, data4robotics} has primarily centered on capturing dynamics in spatiotemporal observations, producing representations effective for imitation learning and robot control.
In parallel, general-purpose visual representation learning has increasingly emphasized visual-language alignment for broad transfer.
From this perspective, conventional robot models remain limited, being largely optimized for narrow control distributions rather than language-grounded semantics.

Learning robot representations directly from paired video-language data offers a direct route to unifying embodied dynamics with language-level semantics.
However, collecting robot vision-action data at scale~\citep{cortexbench,ego4d,data4robotics} is expensive and typically limited to laboratory settings, making it difficult to cover diverse scenes, objects, and interactions.
Therefore, prior work has turned to large-scale video data without robot actions to capture physical dynamics that may benefit downstream robot learning.
While such approaches improve scalability and provide temporal supervision, they still lack the broad semantic coverage and vision-language alignment available in foundation models trained on web-scale corpora.
This motivates a different strategy that adapts already well-aligned foundation models~\citep{clip, siglip, siglip2, theia, oquab2023dinov2} to the requirements of robot control.

In this paper, we propose \textbf{SPARK} (Simple Post-training for Adapting pRetrained Knowledge), a post-training method that adapts foundation models and general-purpose vision models to robot control.
The central goal of SPARK is to make pretrained representations more suitable for control without sacrificing the general visual knowledge acquired during large-scale pretraining.
To this end, SPARK is built on two key principles: \textbf{temporal abstraction} and \textbf{knowledge preservation}.
Temporal abstraction encourages the encoder to produce compact features that preserve information essential for understanding temporal changes by leveraging the prior knowledge acquired during large-scale pretraining.
However, naively enforcing such compact abstraction can degrade useful prior knowledge inherited from large-scale pretraining. Thus, we adopt a retention objective that aligns the patch-wise representations of the adapted encoder with those of the original foundation model.

We empirically demonstrate that SPARK yields consistent improvements across multiple robotics benchmarks~\citep{dmc,metaworld,franka_kitchen, liu2023libero}.
Models adapted with SPARK yield higher success rates and better generalization than their vanilla models on average on robot manipulation tasks, highlighting that cost-efficient post-training can substantially facilitate the generalizability of the foundation model on robot control.
Moreover, experiments on real-world robot tasks show that these gains extend beyond simulation.
Consequently, SPARK presents a promising recipe for mitigating the mismatch between foundation models and the visual representations required for embodied intelligence.

%% file: sec/2_related_work.tex
\section{Related Work}

\paragraph{Pretrained Vision and Foundation Models.}
Large-scale vision and vision-language models, including the CLIP and SigLIP families~\citep{clip, siglip, siglip2} as well as self-supervised models such as DINO, DINOv2, MAE, SimCLR, and MoCo variants~\citep{dino, oquab2023dinov2, mae, simclr, moco, mocov2, mocov3}, provide strong general-purpose visual features and have become standard backbones for many vision tasks.
However, their objectives target recognition and language alignment rather than the temporally fine-grained structure crucial for control, motivating post-training recipes that adapt these models to robotic perception without pre-training from scratch.

\paragraph{Visual Representations in Robotics.}
Pretrained vision backbones are widely used to encode observations for downstream manipulation, navigation, and policy learning~\citep{data4robotics, rlbench}, and even frozen generic features can yield strong representations for imitation learning and control~\citep{parisi22}.
To further improve transfer, robotics-oriented pretraining has exploited egocentric viewpoints and temporal structure through time-contrastive learning with language supervision~\citep{nair2022r3m}, value-implicit pretraining~\citep{ma2022vip}, MAE-style reconstruction on egocentric corpora~\citep{radosavovic2023mvp, cortexbench}, language-guided semantic supervision~\citep{karamcheti2023voltron, zeng2024mpi}, and distillation of diverse foundation models into a single backbone~\citep{theia}.
However, these recipes typically demand substantial computational cost and paired language annotations, limiting their accessibility relative to web-scale foundation models.

\paragraph{Post-training and Adaptation.}
Post-training reduces the cost of adapting pretrained encoders to new domains or tasks.
Parameter-efficient methods introduce small learnable components such as prompts and adapters~\citep{zhou2022coop, zhou2022cocoop, gao2021clip} or apply low-rank updates~\citep{hu2022lora}, while in robotics prior work has used masked-autoencoding~\citep{mae} reconstruction to enhance pretrained encoders for control~\citep{cortexbench}, STP~\citep{stp} learns content and motion features through spatiotemporal pixel prediction, and Lossless Adaptation~\citep{sharma2023lossless} inserts parameter-efficient adapters to preserve the original representation.
Our approach follows this spirit but targets the representation more directly. SPARK introduces a temporal abstraction loss that trains a single class token as a compact state, together with a knowledge preservation loss that aligns token-level features with those of the original foundation model.

\input{figures/framework_v2}

%% file: figures/framework_v2.tex
\begin{figure*}[t]
    \centering
    \vspace{0.5em}
    \includegraphics[width=\linewidth]{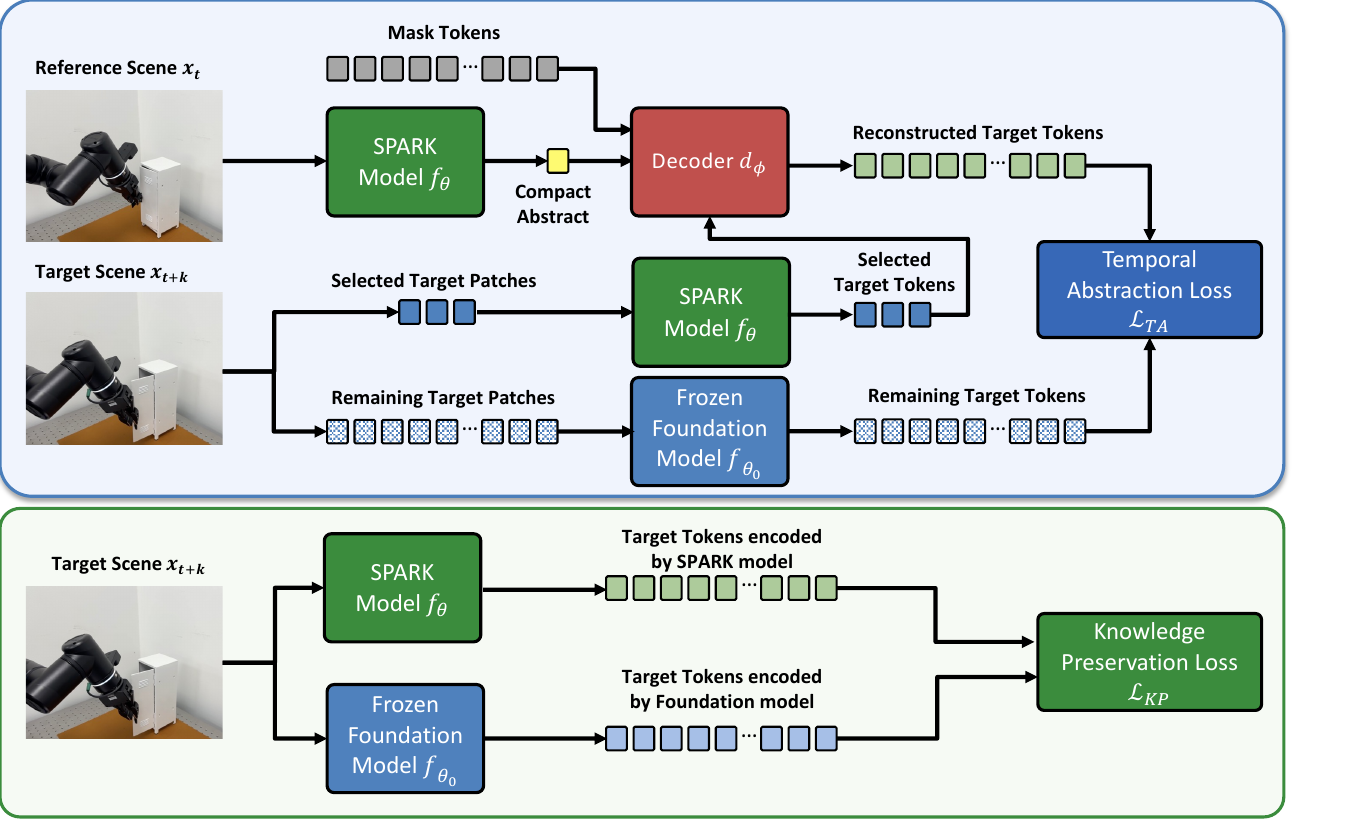}
    \caption{\textbf{Overview of SPARK (Simple Post-training for Adapting pRetrained Knowledge).}
    SPARK encourages temporal abstraction of patch-wise semantics while preserving the foundation model's pre-trained knowledge.
    Given a reference scene $x_t$, the SPARK encoder $f_{\theta}$ produces a class token as a compact abstraction of $x_t$, from which the decoder predicts target token representations with hints from a few target patches.
    The Temporal Abstraction Loss $\mathcal{L}_{\mathrm{TA}}$ aligns these predictions with the target representations of a frozen foundation model $f_{\theta_0}$, while the Knowledge Preservation Loss $\mathcal{L}_{\mathrm{KP}}$ aligns the target patch tokens of $f_{\theta}$ and $f_{\theta_0}$.
    This encourages the class token to serve as a compact visual state for control while retaining the patch-level semantics of the original foundation model.
    }
    \label{fig:method}
\end{figure*}

%% file: sec/3_method.tex
\section{Method}

\subsection{Overview}
The goal of SPARK (\textbf{S}imple \textbf{P}ost-training for \textbf{A}dapting p\textbf{R}etrained \textbf{K}nowledge) is to adapt pretrained foundation models to robot control without sacrificing the broad visual knowledge acquired during large-scale pretraining. To this end, SPARK is built on two complementary principles: \textbf{temporal abstraction} and \textbf{knowledge preservation}. Temporal abstraction encourages the encoder to derive compact features that preserve information essential for understanding temporal changes and action-relevant scene structure. Knowledge preservation regularizes this adaptation so that the learned representation remains aligned with the semantically meaningful features of the original foundation model. Together, these principles enable pretrained representations to become more suitable for robot control without requiring training from scratch.

To realize temporal abstraction, SPARK trains the encoder so that a compact feature extracted from a reference observation can support the prediction of a temporally offset target scene. Specifically, the encoder processes a reference scene and produces patch tokens together with a class token, where the class token serves as a compact feature. A decoder then reconstructs token representations of the target scene from this compact feature, conditioned on only a small subset of visible target patches as hints. Because accurate prediction requires information that explains how the scene changes over time, this objective encourages the compact feature to preserve temporal and action-relevant structure rather than only static global semantics.

However, adapting the encoder solely through this abstraction objective can distort useful pretrained knowledge. To mitigate this issue, SPARK introduces a retention objective that aligns the patch-wise representations of the adapted encoder with those of the frozen original foundation model on the same input. This regularization preserves semantically meaningful local features while allowing the compact feature to specialize toward dynamic understanding for control. The overall pipeline is illustrated in Fig.~\ref{fig:method}.

\subsection{Formulation}

\paragraph{Encoding procedure.}
Let $f_{\theta}$ denote the SPARK encoder and let $f_{\theta_0}$ denote the frozen foundation model used for initialization. We sample a reference scene $x_t \in \mathbb{R}^{3 \times H \times W}$ and a temporally offset target scene $x_{t+k} \in \mathbb{R}^{3 \times H \times W}$. We patchify both scenes into $N$ non-overlapping patches, denoted by $\{x_t^i\}_{i=1}^N$ and $\{x_{t+k}^i\}_{i=1}^N$, respectively.
The reference scene is processed by both the SPARK encoder $f_{\theta}$ and the frozen encoder $f_{\theta_0}$, yielding adapted token representations $\{u_t^i\}_{i=1}^N$ and frozen token representations $\{v_t^i\}_{i=1}^N$. In addition, the SPARK encoder produces a class token $u_t^{[\mathrm{CLS}]}$, which serves as the compact feature associated with the reference scene. For the target scene, the frozen encoder processes all target patches to produce $\{v_{t+k}^i\}_{i=1}^N$, while the SPARK encoder processes only a subset of visible target patches indexed by $M \subset \{1,\dots, N\}$ during temporal abstraction, where $|M|=\lfloor rN \rfloor$ for a visible ratio $r \in (0,1)$. This yields adapted target tokens $\{u_{t+k}^i\}_{i \in M}$, which are used as hints for target-scene prediction. For knowledge preservation, the SPARK encoder additionally processes all target patches to produce $\{u^i_{t+k}\}_{i=1}^N$.

\paragraph{Temporal abstraction loss.}
The abstraction objective encourages the compact feature $u_t^{[\mathrm{CLS}]}$ to retain the information required to predict the target scene. To this end, we reconstruct target token representations from the compact feature together with the visible target hints $\{u_{t+k}^i\}_{i \in M}$. Let $M^c$ denote the masked target-patch indices. We introduce learned mask tokens $\{m_i\}_{i \in M^c}$ and feed the concatenation of the compact feature, visible target tokens, and mask tokens into a decoder $d_{\phi}$. The decoder predicts target token representations $\{\hat{u}_{t+k}^i\}_{i \in M^c}$ for the masked regions.
We then align these predicted target tokens with the corresponding frozen target representations produced by the original foundation model:
\begin{equation}
\mathcal{L}_{\mathrm{TA}}
=
\sum_{i \in M^c}
d\!\left(\hat{u}_{t+k}^i,\; v_{t+k}^i\right),
\end{equation}
where $d(\cdot,\cdot)$ is a patch-level distance function. By requiring the compact feature to support prediction of the target scene, this loss encourages the encoder to abstract temporal and action-relevant information from the reference observation into a compact representation.

\paragraph{Knowledge preservation loss.}
While the abstraction loss specializes the representation toward temporal understanding, it may also distort the useful semantic structure inherited from the pretrained model. To prevent this drift, we introduce a retention loss that aligns the patch-wise representations of the adapted encoder with those of the frozen original encoder on the target scene:
\begin{equation}
\mathcal{L}_{\mathrm{KP}}
=
\sum_{i=1}^{N}
d'\!\left(u_{t+k}^i,\; v_{t+k}^i\right),
\end{equation}
where $d'(\cdot,\cdot)$ is a patch-level distance function. Since $f_{\theta_0}$ is frozen, the resulting representations provide a stable semantic target during post-training. This regularization preserves general visual knowledge while allowing the adapted encoder to specialize in robot control.

\paragraph{Post-training objective.}
The overall SPARK objective combines temporal abstraction with knowledge preservation:
\begin{equation}
\mathcal{L}_{\mathrm{SPARK}}
=
\mathcal{L}_{\mathrm{TA}}
+
\alpha \mathcal{L}_{\mathrm{KP}},
\end{equation}
where $\alpha > 0$ balances adaptation and preservation. Optimizing this objective yields compact visual features that capture temporal and action-relevant information while remaining grounded in the semantic knowledge of the original foundation model.

\input{tables/quan_comp_avg}

%% file: tables/quan_comp_avg.tex
\begin{table*}[t]
\caption{\textbf{Experimental results on vision-based robot policy learning across MetaWorld, DMC, and Franka Kitchen.} We report the average performance of imitation learning agents trained upon representations of CLIP~\citep{clip}, DINOv2~\citep{oquab2023dinov2}, SigLIP~\citep{siglip}, and Theia~\citep{theia} post-trained on Kinetics-400~\citep{kay2017kinetics} dataset. Success rates (\%) are reported for MetaWorld~\citep{metaworld} and Franka Kitchen~\citep{franka_kitchen}, and normalized scores are reported for DMC~\citep{dmc}. We report the gains of our method over the baseline.}
\label{tab:combined_avg}
\centering
\tabcolsep=.85em
\begin{tabular}{l|l|ccc|c}
\toprule
\multicolumn{2}{c|}{Method} & MetaWorld & DMC & Franka Kitchen & Average \\
\midrule
\multirow{3}{*}{CLIP}   & Baseline & 72.8{\scriptsize $\pm$1.1} & 47.9{\scriptsize $\pm$2.1} & 31.5{\scriptsize $\pm$2.4} & 50.7 \\
                        & SPARK    & 77.7{\scriptsize $\pm$7.0} & 50.6{\scriptsize $\pm$0.9} & 39.8{\scriptsize $\pm$5.6} & 56.0 \\
                        & Gain     & \textcolor{darkergreen}{+ 4.9} & \textcolor{darkergreen}{+ 2.7} & \textcolor{darkergreen}{+ 8.3} & \textcolor{darkergreen}{+ 5.3} \\
\midrule
\multirow{3}{*}{SigLIP} & Baseline & 73.0{\scriptsize $\pm$7.6} & 42.1{\scriptsize $\pm$1.5} & 31.2{\scriptsize $\pm$0.6} & 48.8 \\
                        & SPARK    & 76.9{\scriptsize $\pm$6.8} & 42.6{\scriptsize $\pm$1.8} & 42.2{\scriptsize $\pm$2.4} & 53.9 \\
                        & Gain     & \textcolor{darkergreen}{+ 3.9} & \textcolor{darkergreen}{+ 0.5} & \textcolor{darkergreen}{+ 11.0} & \textcolor{darkergreen}{+ 5.1} \\
\midrule
\multirow{3}{*}{DINOv2} & Baseline & 78.4{\scriptsize $\pm$7.7} & 55.3{\scriptsize $\pm$0.4} & 36.4{\scriptsize $\pm$2.7} & 56.7 \\
                        & SPARK    & 83.2{\scriptsize $\pm$5.5} & 58.3{\scriptsize $\pm$2.4} & 40.7{\scriptsize $\pm$2.7} & 60.7 \\
                        & Gain     & \textcolor{darkergreen}{+ 4.8} & \textcolor{darkergreen}{+ 3.0} & \textcolor{darkergreen}{+ 4.3} & \textcolor{darkergreen}{+ 4.0} \\
\midrule
\multirow{3}{*}{Theia}  & Baseline & 89.0{\scriptsize $\pm$2.8} & 60.8{\scriptsize $\pm$2.8} & 40.1{\scriptsize $\pm$2.3} & 63.3 \\
                        & SPARK    & 89.2{\scriptsize $\pm$7.4} & 69.3{\scriptsize $\pm$1.4} & 44.3{\scriptsize $\pm$2.6} & 67.6 \\
                        & Gain     & \textcolor{darkergreen}{+ 0.2} & \textcolor{darkergreen}{+ 8.5} & \textcolor{darkergreen}{+ 4.2} & \textcolor{darkergreen}{+ 4.3} \\
\bottomrule
\end{tabular}
\end{table*}

%% file: sec/4_experiments.tex
\section{Experiments}
In this section, we evaluate the impact of the proposed method in vision-based policy learning for robotic manipulation across various environments~\citep{franka_kitchen, metaworld, dmc}, and validate that such improvements transfer to real-world demonstrations.
Moreover, we compare our method against various robot representation learning methods, with additional comparisons to commonly used adaptation methods provided in the Appendix.

\input{figures/qual_real}
\input{tables/quan_real}

\subsection{Vision-based Robot Policy Learning in Simulated Environments}
\label{sec:eval_il_sim}
We evaluate our method through imitation learning on robot manipulation and locomotion tasks across various simulated environments.
Specifically, we evaluate on five tasks from each of the Franka Kitchen, MetaWorld, and DeepMind Control Suite (DMC)~\citep{dmc} benchmarks; details for the benchmarks and tasks are deferred to the Appendix.
We compare our method against baselines built upon four different visual representations: CLIP~\citep{clip}, SigLIP~\citep{siglip}, DINOv2~\citep{oquab2023dinov2}, and Theia~\citep{theia}, all post-trained on the Kinetics-400~\citep{kay2017kinetics} dataset.
Table~\ref{tab:combined_avg} summarizes the average performance of our method (SPARK) and the baselines across the three benchmarks.
The results demonstrate that our method consistently outperforms the baselines across all three benchmarks and all four visual representations.
In particular, SPARK improves the average performance by 5.3$\%$p, 5.1$\%$p, and 4.0$\%$p over the CLIP, SigLIP, and DINOv2 baselines, respectively, and further boosts the performance by 8.5$\%$p on DMC when built upon Theia, reaching $69.3\%$ normalized score on DMC and $89.2\%$ success rate on MetaWorld.
Notably, our method achieves the largest gains on the Franka Kitchen benchmark, with average success rate improvements of 8.3$\%$p, 11.0$\%$p, 4.3$\%$p and 4.2$\%$p on CLIP, SigLIP, DINOv2, and Theia, respectively, suggesting that our method is particularly effective when accurate visual state grounding is critical for fine-grained manipulation.
This highlights the effectiveness of explicitly encoding visual state representation for vision-based robot policy learning, regardless of the underlying visual backbone.

\subsection{Vision-based Robot Policy Learning in Real-world Environments}
\paragraph{Quantitative results.}
To evaluate whether the benefits of SPARK extend beyond simulation, we additionally conduct experiments on real-world robot manipulation tasks. We consider three imitation learning tasks: \textit{Cloth Folding}, \textit{Cup Stacking}, and \textit{Cabinet Opening}. For each task, we collect 20 demonstration episodes for behavior cloning and evaluate policies learned from visual representations extracted by the frozen backbone. The robot operates at 20 Hz with a policy's chunk size of 20.
Following the same evaluation protocol as in the simulation, we compare the vanilla CLIP encoder with its SPARK-adapted counterpart with 20 trials for each task. The results are reported in Table~\ref{tab:rw}.
SPARK improves the success rate on all three tasks, increasing performance from 65.0 to 75.0 on \textit{Cloth Folding}, from 45.0 to 55.0 on \textit{Cup Stacking}, and from 50.0 to 55.0 on \textit{Cabinet Opening}. Overall, these results show that the gains obtained by SPARK are not limited to simulation, but are transferable to real-world robot manipulation as well. This suggests that post-training foundation models with temporal abstraction and knowledge preservation improves the usefulness of pretrained visual representations in practical robotic settings.

\paragraph{Qualitative results.}
We further visualize representative real-world rollouts of the SPARK-adapted policy across multiple manipulation tasks in Fig.~\ref{fig:qual_real}. Each row presents a temporal sequence from the initial scene, through intermediate interaction stages, to the final task outcome. The visualizations show that the policy maintains coherent manipulation progress throughout the rollout, including approach, contact, and task completion. Across different tasks and scene configurations, these examples suggest that SPARK produces visual representations that preserve the information most relevant for control, enabling robust real-world execution under variations in object pose, scene layout, and appearance.

\input{tables/comp_models}

\subsection{Comparison to prior pretraining methods}
We compare SPARK against representative robot-oriented representation learning methods on the MetaWorld benchmark~\citep{metaworld}, following the same imitation learning protocol as in Sec.~\ref{sec:eval_il_sim}.
The baselines are grouped into two categories: \textit{self-supervised learning} methods including R3M~\citep{nair2022r3m}, VC-1~\citep{cortexbench}, and data4robotics~\citep{data4robotics}, and \textit{methods with auxiliary language guidance} including MVP~\citep{radosavovic2023mvp}, Voltron~\citep{karamcheti2023voltron}, and MPI~\citep{zeng2024mpi}.
As shown in Table~\ref{tab:comp_all}, SPARK-trained Theia attains the best overall performance of $89.2\%$, surpassing all robot-specific baselines from both categories, while SPARK-trained DINOv2 ($83.2\%$) is also competitive.
Overall, adapting public foundation models with SPARK is more effective for downstream robotics tasks than training robot-specific representations from scratch, even when the latter benefit from auxiliary language supervision.

%% file: figures/qual_real.tex
\begin{figure*}[t]
    \centering
    \setlength{\tabcolsep}{1pt}
    \renewcommand{\arraystretch}{1.0}
    \begin{tabular}{@{}c@{\hspace{2pt}}cccccc@{}}
        \multicolumn{7}{@{}c@{}}{Cloth Folding} \\[2pt]

        \makebox[5mm][c]{\raisebox{.2\height}{\rotatebox{90}{\small CLIP (SPARK)}}} &
        \includegraphics[width=0.16\textwidth]{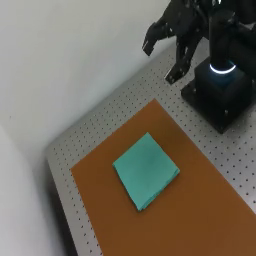} &
        \includegraphics[width=0.16\textwidth]{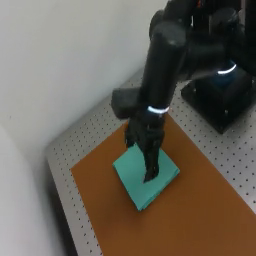} &
        \includegraphics[width=0.16\textwidth]{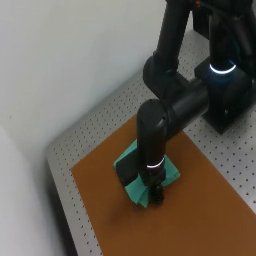} &
        \includegraphics[width=0.16\textwidth]{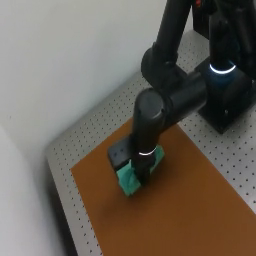} &
        \includegraphics[width=0.16\textwidth]{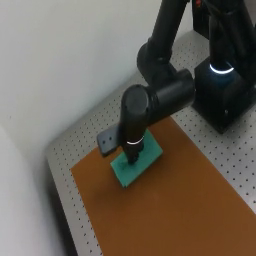} &
        \includegraphics[width=0.16\textwidth]{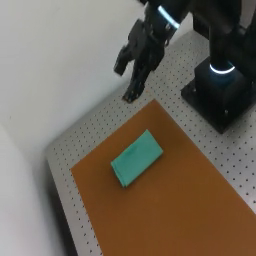} \\

        \makebox[5mm][c]{\raisebox{1.7\height}{\rotatebox{90}{\small CLIP}}} &
        \includegraphics[width=0.16\textwidth]{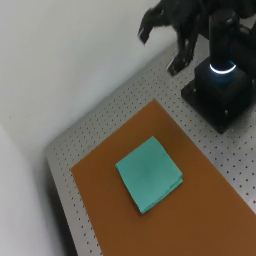} &
        \includegraphics[width=0.16\textwidth]{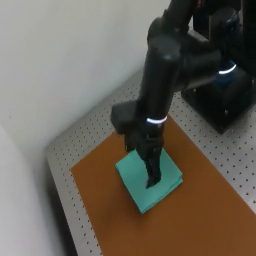} &
        \includegraphics[width=0.16\textwidth]{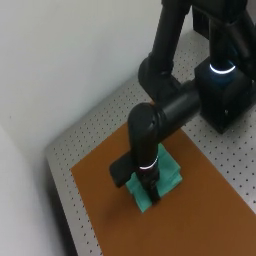} &
        \includegraphics[width=0.16\textwidth]{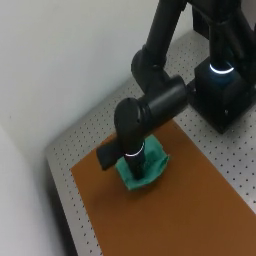} &
        \includegraphics[width=0.16\textwidth]{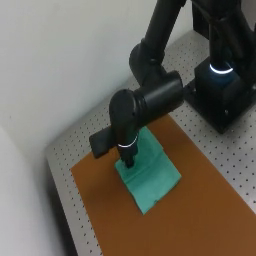} &
        \includegraphics[width=0.16\textwidth]{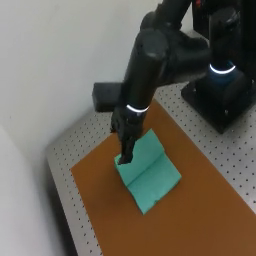} \\

        \multicolumn{7}{@{}c@{}}{Cup stacking} \\[2pt]

        \makebox[5mm][c]{\raisebox{.2\height}{\rotatebox{90}{\small CLIP (SPARK)}}} &
        \includegraphics[width=0.16\textwidth]{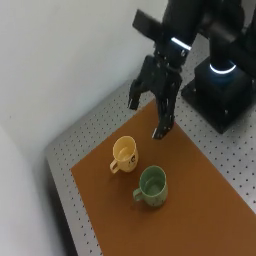} &
        \includegraphics[width=0.16\textwidth]{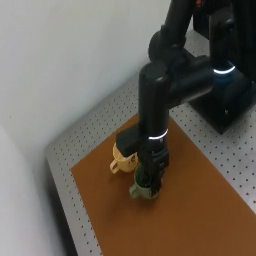} &
        \includegraphics[width=0.16\textwidth]{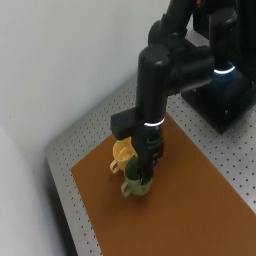} &
        \includegraphics[width=0.16\textwidth]{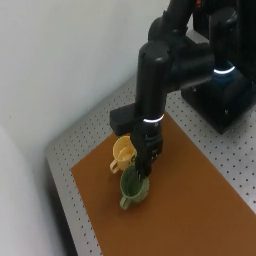} &
        \includegraphics[width=0.16\textwidth]{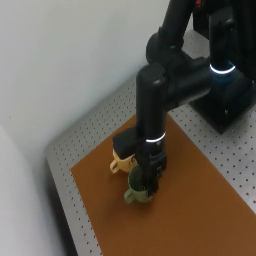} &
        \includegraphics[width=0.16\textwidth]{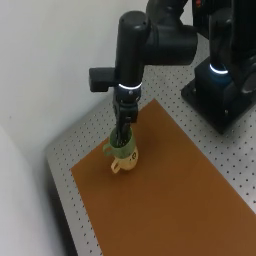} \\

        \makebox[5mm][c]{\raisebox{1.7\height}{\rotatebox{90}{\small CLIP}}} &
        \includegraphics[width=0.16\textwidth]{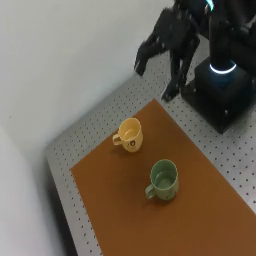}  &
        \includegraphics[width=0.16\textwidth]{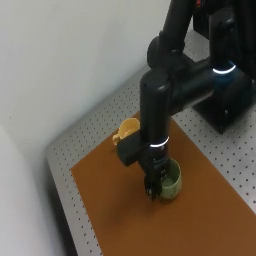} &
        \includegraphics[width=0.16\textwidth]{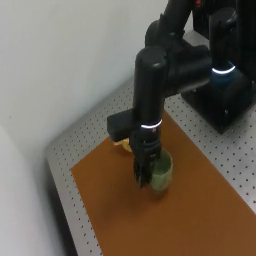} &
        \includegraphics[width=0.16\textwidth]{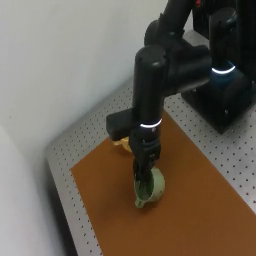} &
        \includegraphics[width=0.16\textwidth]{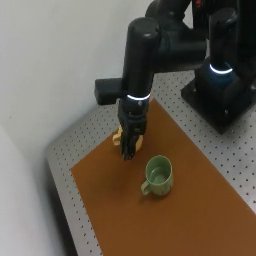} &
        \includegraphics[width=0.16\textwidth]{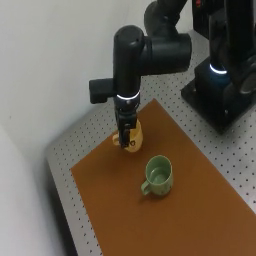}
    \end{tabular}
    \caption{\textbf{Real-world demonstration trajectories.} Each row shows the temporal progression of a policy from the initial observation to intermediate interaction states and the final outcome. Across tasks, the SPARK-adapted policy exhibits stable approach, contact, and object manipulation behaviors, illustrating that the adapted visual representations capture control-relevant scene information that transfers reliably to real-world settings.}
    \label{fig:qual_real}
\end{figure*}

%% file: tables/quan_real.tex
\begin{figure}[t]
    \centering
    \small
    \begin{minipage}[t]{\linewidth}
        \centering
        \tabcolsep=0.7em
        \captionof{table}{\textbf{Performance on real-world vision-based robot policy learning.} Success rates (\%) of imitation learning agents on three manipulation tasks: Cloth Folding, Cup Stacking, and Cabinet Opening. We compared CLIP~\citep{clip} with our SPARK model, post-trained upon representations of CLIP on Kinetics-400~\citep{kay2017kinetics}. The results demonstrate the generalizability of SPARK in the real world.}
        \label{tab:rw}
        \vspace{-.2em}
        \begin{adjustbox}{width=0.5\linewidth}
        \begin{tabular}{lcccc}
            \toprule
            Method & \makecell{Cloth\\Folding} & \makecell{Cup\\Stacking} & \makecell{Cabinet\\Opening}  \\
            \midrule
            CLIP & 65.0 & 45.0 & 50.0   \\
            SPARK (CLIP)      & 75.0   & 55.0   & 55.0 \\
            \bottomrule
        \end{tabular}
        \end{adjustbox}
    \end{minipage}

    \vspace{-1em}
\end{figure}

%% file: tables/comp_models.tex
\begin{wraptable}{r}{0.4\textwidth}
\centering
\small
\tabcolsep=.4em
\vspace{-4em}
\caption{\textbf{Comparative analysis of vision-based robot policy learning on MetaWorld.} We report the performance of imitation learning agents on Metaworld \citep{metaworld}. The success rates (\%) are reported for all the tasks.}
\label{tab:comp_all}
\resizebox{0.9\linewidth}{!}{%
\begin{tabular}{lcc}
\toprule
Methods & Architecture & Metaworld \\
\midrule
\multicolumn{3}{l}{\textit{Self-supervised Learning}} \\
R3M$^\circ$    & ViT-S & 67.0 \\
VC-1           & ViT-B & 86.4 \\
data4robotics  & ViT-B & 87.0 \\
\midrule
\multicolumn{3}{l}{\textit{Supervision with Aux. Language Guidance}} \\
R3M                & ViT-S & 69.2 \\
MVP                & ViT-S & 84.6 \\
Voltron$^\ddagger$ & ViT-S & 68.7 \\
MPI$^\ddagger$     & ViT-S & 85.7 \\
\midrule
\multicolumn{3}{l}{\textit{SPARK Models}} \\
CLIP    & ViT-B & 77.7 \\
SigLIP  & ViT-B & 76.9 \\
DINOv2  & ViT-B & 83.2 \\
Theia   & ViT-B & \textbf{89.2} \\
\bottomrule
\end{tabular}%
}

\vspace{0.3em}
{\scriptsize $^\circ$ Excludes language guidance from the vanilla recipe.}
\vspace{-3em}
\end{wraptable}

%% file: sec/5_analysis.tex
\input{tables/vla-libero}

\section{Analysis}
\label{sec:analysis}

\paragraph{SPARK as VLA vision tower.}
We examine whether SPARK representations also benefit vision-language-action (VLA) systems.
We construct a VLA by attaching a randomly initialized action expert to a VLM with either vanilla SigLIP or SPARK-SigLIP as its vision tower and SmolLM2-360M~\citep{allal2025smollm2} as its LLM decoder, and train the action expert from scratch on LIBERO-10~\citep{liu2023libero} with the entire VLM frozen.
We consider two settings: an aligned VLM, where projector alignment is conducted on LLaVA-Pretrain~\citep{liu2023llava} beforehand, and a \textit{non-aligned VLM}, which skips this step.
As shown in Table~\ref{tab:vla-libero}, SPARK-SigLIP consistently outperforms vanilla SigLIP in both settings.
While projector alignment remains beneficial for absolute performance, the consistent SPARK gain across both settings indicates that its dynamic and action-relevant structure transfers to LLM-conditioned policy learning.

\input{tables/vl-score}

\paragraph{Vision-language alignment is preserved after SPARK post-training.}
A natural concern with any post-training procedure is that it may distort the foundation model's original vision-language alignment, since the encoder is updated on a video corpus without any paired captions.
To probe this directly, we compare image-text retrieval performance of SPARK-trained SigLIP against vanilla SigLIP on Flickr30K-1K~\citep{young2014image} test, which contains 5,000 image-text pairs.
We employ the original SigLIP text encoder for text token encoding.
We report Recall@$K$ (R@$K$), which measures the fraction of queries for which the correct match appears among the top-$K$ retrieved candidates, with $K \in \{1, 5, 10\}$.
As reported in Table~\ref{tab:siglip-retention}, SPARK retains nearly all of vanilla SigLIP's retrieval ability across both directions and all values of $K$.
Specifically,
the largest drop is -2.8$\%$p at Image$\to$Text R@1, while R@5 and R@10 are within one point of vanilla in both directions.
Despite reconstruction-driven post-training on video data, the vision encoder remains compatible with the original text encoder it was paired with at pretraining time.
This indicates that our Knowledge Preservation loss successfully prevents the post-trained representation from drifting away from the language-aligned manifold of the underlying vision-language model.

\begin{table*}[t]
\caption{\textbf{Per-loss ablation.} We ablate each loss individually using SigLIP on Franka Kitchen. The success rates (\%) are reported for all the tasks.}
\label{tab:per-loss}
\centering
\small
\tabcolsep=.5em
\begin{tabular}{l|ccccc|c}
\toprule
Variant & Knob1 on & Light on & Sdoor open & Ldoor open & Micro open & Average \\
\midrule
SigLIP & 17.0{\scriptsize $\pm$1.2} & 38.5{\scriptsize $\pm$1.9} & 75.5{\scriptsize $\pm$3.4} & 8.5{\scriptsize $\pm$1.0} & 16.5{\scriptsize $\pm$1.9} & 31.2{\scriptsize $\pm$0.6} \\
$\mathcal{L}_{\mathrm{TA}}$-only & 5.5{\scriptsize $\pm$1.9} & 11.5{\scriptsize $\pm$5.3} & 43.0{\scriptsize $\pm$6.6} & 6.5{\scriptsize $\pm$1.0} & 6.5{\scriptsize $\pm$1.0} & 14.6{\scriptsize $\pm$2.5} \\
$\mathcal{L}_{\mathrm{KP}}$-only & 12.0{\scriptsize $\pm$2.8} & 19.5{\scriptsize $\pm$1.9} & 67.0{\scriptsize $\pm$2.6} & 7.0{\scriptsize $\pm$1.2} & 8.5{\scriptsize $\pm$1.9} & 22.8{\scriptsize $\pm$0.7} \\
\midrule
SPARK & \textbf{27.5}{\scriptsize $\pm$5.0} & \textbf{43.0}{\scriptsize $\pm$5.0} & \textbf{87.5}{\scriptsize $\pm$1.9} & \textbf{20.0}{\scriptsize $\pm$2.8} & \textbf{33.0}{\scriptsize $\pm$5.3} & \textbf{42.2}{\scriptsize $\pm$2.4} \\
\bottomrule
\end{tabular}
\end{table*}

\paragraph{Ablation study on losses.}
We ablate each loss individually using SigLIP on Franka Kitchen.
As shown in Table~\ref{tab:per-loss}, $\mathcal{L}_{\mathrm{TA}}$-only collapses performance across all five tasks, as the strong abstraction pressure has no semantic anchor.
$\mathcal{L}_{\mathrm{KP}}$-only anchors the representation and thus mitigates the degradation to some extent, but fails to improve control.
Neither loss alone maintains the baseline, and only their combination surpasses it, validating that the two losses are mutually necessary components.

\paragraph{Comparison to various post-training recipes.}
\input{tables/comp_post}

We conduct comparative analysis with alternative post-training recipes on Franka Kitchen. We compared naive post-training using masked autoencoding (MAE)~\citep{mae, cortexbench} and correspondence matching among consecutive scenes~\citep{siammae}. All the post-trained models are initialized by CLIP~\citep{clip}.
As shown in Table~\ref{tab:comp_post}, standard MAE style post-training and correspondence matching style post-training, which focus on generic reconstruction and temporal correspondence respectively, perform worse than the baseline with averages of 23.0\(\%\) and \(25.5\%\). In contrast, SPARK attains an average success rate of \(39.8\%\), substantially outperforming both the vanilla CLIP and other post-trained models.
This highlights that explicitly enforcing temporal abstraction and knowledge preservation is more effective for adapting foundation models to robot control than generic reconstruction based objectives.

\begin{table*}[t]
\caption{\textbf{Ablation study on masking strategies.} We replace random masking with structured variants based on the object-relevance criterion following AutoMAE~\citep{chen2023automae}, using SigLIP on Franka Kitchen. The success rates (\%) are reported for all the tasks.}
\label{tab:masking}
\centering
\small
\tabcolsep=.5em
\begin{tabular}{lcccccc|c}
\toprule
Method & Masking strategy & Knob1 on & Light on & Sdoor open & Ldoor open & Micro open & Average \\
\midrule
SigLIP & - & 17.0{\scriptsize $\pm$1.2} & 38.5{\scriptsize $\pm$1.9} & 75.5{\scriptsize $\pm$3.4} & 8.5{\scriptsize $\pm$1.0} & 16.5{\scriptsize $\pm$1.9} & 31.2{\scriptsize $\pm$0.6} \\
\midrule
\multirow{3}{*}{SPARK} & Object-aware masking & 8.5{\scriptsize $\pm$1.0} & 20.0{\scriptsize $\pm$3.3} & 60.0{\scriptsize $\pm$5.4} & 8.5{\scriptsize $\pm$1.0} & 10.0{\scriptsize $\pm$1.6} & 21.4{\scriptsize $\pm$1.2} \\
& Object-preserving masking & 12.5{\scriptsize $\pm$2.5} & 20.5{\scriptsize $\pm$6.8} & 72.5{\scriptsize $\pm$4.4} & 9.5{\scriptsize $\pm$1.9} & 10.5{\scriptsize $\pm$1.0} & 25.1{\scriptsize $\pm$2.1} \\
& Random masking & \textbf{27.5}{\scriptsize $\pm$5.0} & \textbf{43.0}{\scriptsize $\pm$5.0} & \textbf{87.5}{\scriptsize $\pm$1.9} & \textbf{20.0}{\scriptsize $\pm$2.8} & \textbf{33.0}{\scriptsize $\pm$5.3} & \textbf{42.2}{\scriptsize $\pm$2.4} \\
\bottomrule
\end{tabular}
\end{table*}

\paragraph{Ablation study on masking strategy.}
We replace random masking with two structured variants based on the object-relevance criterion using [CLS] attention following AutoMAE~\citep{chen2023automae}, which either mask all object-relevant patches (i.e., object-aware masking) or always keep them as visible hints (i.e., object-preserving masking).
As shown in Table~\ref{tab:masking}, both variants fall far below random masking.
This is because masking all object-relevant patches leaves only uninformative hints, breaking the reconstruction signal, while always keeping them lets the decoder bypass the class token.
Random masking balances these issues, providing diverse coverage that forces the class token to capture scene-level temporal structure.

\begin{table*}[t]
\caption{\textbf{Generalizability across post-training datasets.} We post-train CLIP on Something-Something-V2 (SSv2), an egocentric hand-object interaction dataset, with the same configuration as Kinetics-400. The success rates (\%) are reported for all the tasks on Franka Kitchen.}
\label{tab:ssv2}
\centering
\small
\tabcolsep=.5em
\begin{tabular}{lcccccc|c}
\toprule
Method & Dataset & Knob1 on & Light on & Sdoor open & Ldoor open & Micro open & Average \\
\midrule
CLIP & - & 23.0{\scriptsize $\pm$2.6} & 29.5{\scriptsize $\pm$4.7} & 69.5{\scriptsize $\pm$4.4} & 13.5{\scriptsize $\pm$3.4} & 22.0{\scriptsize $\pm$2.8} & 31.5{\scriptsize $\pm$2.4} \\
\midrule
\multirow{2}{*}{SPARK}
 & SSv2 & 28.5{\scriptsize $\pm$1.9} & 37.0{\scriptsize $\pm$6.8} & \textbf{78.0}{\scriptsize $\pm$3.7} & \textbf{19.0}{\scriptsize $\pm$4.2} & 15.0{\scriptsize $\pm$3.8} & 35.5{\scriptsize $\pm$1.1} \\
& Kinetics-400 & \textbf{31.5}{\scriptsize $\pm$1.9} & \textbf{38.5}{\scriptsize $\pm$6.0} & 74.5{\scriptsize $\pm$11.4} & 18.0{\scriptsize $\pm$2.0} & \textbf{29.0}{\scriptsize $\pm$6.2} & \textbf{39.8}{\scriptsize $\pm$5.6} \\
\bottomrule
\end{tabular}
\end{table*}

\paragraph{Generalizability across post-training datasets.}
To test whether SPARK depends on third-person video, we post-train CLIP on Something-Something-V2 (SSv2)~\citep{ssv2}, an egocentric hand-object interaction dataset, under the same configuration.
SSv2-trained SPARK improves the baseline on most of the tasks and by 4.0$\%$p on average, confirming that SPARK transfers to egocentric videos.
The gain is smaller than that of Kinetics-400, likely due to narrower scene diversity of SSv2.

%% file: tables/vla-libero.tex
\begin{wraptable}{r}{0.45\textwidth}
\centering
\small
\vspace{-1em}
\caption{\textbf{Comparative analysis of vision towers on VLA systems for long-horizon tasks.} We consider two settings: an aligned VLM setting, where projector alignment is first conducted on the LLaVA-Pretrain dataset of 558K image-caption pairs while the non-aligned VLM setting skips this step. In both settings, the action expert is trained from scratch on LIBERO-10 while the full VLM remains frozen. Success rates are reported over 500 rollouts with 50 rollouts per task.}

\label{tab:vla-libero}
\resizebox{\linewidth}{!}{%
\begin{tabular}{lcc}
\toprule
\multirow{2}{*}{Method} & \multicolumn{2}{c}{Success rate ($\%$)}  \\
 & Aligned VLM & Non-aligned VLM\\
\midrule
SigLIP         & 39.2  & 29.5\\
SPARK (SigLIP) & 41.4 & 31.2\\
\bottomrule
\end{tabular}%
}
\vspace{-1em}
\end{wraptable}

%% file: tables/vl-score.tex
\begin{table}[t]
\centering
\caption{\textbf{SPARK preserves vision-language alignment.} Image-Text and Text-Image retrieval on Flickr30K 1K test~\citep{young2014image}, which consists of  5,000 ground-truth pairs. We use the original SigLIP text encoder, so any drop is attributable to vision-side drift. $\Delta$ denotes the difference between the SPARK model and its baseline.}
\label{tab:siglip-retention}
\small
\begin{tabular}{lrrrrrr}
\toprule
 & \multicolumn{3}{c}{Image $\to$ Text} & \multicolumn{3}{c}{Text $\to$ Image} \\
\cmidrule(lr){2-4} \cmidrule(lr){5-7}
Model & R@1 & R@5 & R@10 & R@1 & R@5 & R@10 \\
\midrule
SigLIP         & 90.4 & 98.6 & 99.0 & 75.6 & 93.2 & 96.3 \\
SPARK (SigLIP) & 87.6 & 97.7 & 98.8 & 74.0 & 92.4 & 95.6 \\
\midrule
$\Delta$              & $-2.8$ & $-0.9$ & $-0.2$ & $-1.6$ & $-0.8$ & $-0.7$ \\
\bottomrule
\end{tabular}
\end{table}

%% file: tables/comp_post.tex
\begin{wraptable}{r}{0.4\textwidth}
\centering
\small
\vspace{-1.2em}
\caption{\textbf{Ablation study varying post-training recipes.} We compared our SPARK to other post-training recipes on Franka Kitchen~\citep{franka_kitchen} using CLIP~\citep{clip} as an initial model. The success rates (\%) are reported for all the tasks.
}
\label{tab:comp_post}
\resizebox{\linewidth}{!}{%
\begin{tabular}{lc}
\toprule
Post-training & Franka Kitchen \\
\midrule
Baseline                & 31.5 \\
Masked autoencoding     & 23.0 \\
Correspondence matching & 25.5 \\
\midrule
SPARK                   & \textbf{39.8} \\
\bottomrule
\end{tabular}%
}
\vspace{-1em}
\end{wraptable}

%% file: sec/6_conclusion.tex
\section{Discussion}
\label{sec:conclusion}

\paragraph{Summary.}
In this paper, we proposed SPARK, a simple post-training method for adapting pretrained foundation models to robot control.
SPARK is based on two key principles: temporal abstraction and knowledge preservation.
Temporal abstraction encourages the learned representation to capture temporal changes relevant to control, while knowledge preservation prevents the adapted model from losing useful semantic knowledge inherited from large-scale pre-training.
In this way, SPARK makes pretrained visual representations more suitable for robot control without requiring training from scratch.
Experiments on multiple robot learning benchmarks show that SPARK consistently improves over vanilla pretrained backbones.
We further show that these gains extend beyond simulation to real-world robot manipulation tasks.
These results suggest that cost-efficient post-training is a practical way to improve the usefulness of foundation models for embodied intelligence.

\paragraph{Limitations and Future Work.}
While SPARK consistently improves over vanilla foundation models across multiple benchmarks, several limitations remain.
First, although SPARK consistently improves each backbone over its baselines, the absolute downstream performance still depends substantially on the capacity of the underlying foundation model, indicating that post-training amplifies rather than replaces the capacity of the base representation.
Second, our temporal abstraction is built on a relatively simple temporal formulation that predicts a single temporally offset scene from a reference observation.
To address these limitations, promising directions for future work include extending the temporal formulation beyond a single offset scene, such as multi-step prediction, and integrating SPARK into full VLA systems~\citep{kim2024openvla, intelligence2025pi, zheng2026xvla, bjorck2025groot}, where the vision tower is trained jointly with the policy.

%% file: sec/x_appendix.tex
\newpage

\appendix

\section*{Appendix}

\setcounter{table}{0}
\renewcommand{\thetable}{\Alph{table}}
\setcounter{figure}{0}
\renewcommand{\thefigure}{\Alph{figure}}
\setcounter{section}{0}
\renewcommand\thesection{\Alph{section}}

\begin{itemize}
    \item \S\ref{sec:eval}: Evaluation suites
    \item \S\ref{sec:detailed_quan_sim}: Per-task Results of Vision-based Robot Policy Learning in Simulated Environments
    \item \S\ref{sec:adaptation}: Comparison to Representation Adaptation Methods
    \item \S\ref{sec:imple}: Implementation details
\end{itemize}

\section{Evaluation suites} \label{sec:eval}
We evaluate on three vision-based robot learning benchmarks encompassing Franka Kitchen, DeepMind Control Suite, and MetaWorld, covering a total of 15 tasks.
The environments and sampled tasks for each benchmark are shown in Fig.~\ref{fig:benchmarks}.\\

\noindent\textbf{DeepMind Control Suite}~\citep{dmc} is a set of continuous control tasks with simulated robots.
We use five imitation learning cases: \textit{Walker-stand}, \textit{Walker-walk}, \textit{Reacher-easy}, \textit{Cheetah-run}, and \textit{Finger-spin}. We report the normalized scores for all tasks.\\

\noindent\textbf{MetaWorld}~\citep{metaworld} is a suite of simulated robotic manipulation tasks with a Sawyer robot arm. We focus on a subset of five representative tasks: \textit{Assembly}, \textit{Bin-picking}, \textit{Button-press-topdown}, \textit{Drawer-open}, and \textit{Hammer}. We measure the best success rates among the online evaluation trials.

\noindent\textbf{Franka Kitchen}~\citep{franka_kitchen}
is a challenging robotic manipulation environment, which consists of a 9 DoF position-controlled Franka robot interacting with a kitchen scene. The environment contains various kitchenwares including an openable microwave, four turnable oven burners, an oven light switch, two hinged cabinets, and a sliding cabinet door. We consider the imitation learning evaluation setup with five imitation tasks: \textit{Knob1 on}, \textit{Light on}, \textit{Sdoor open}, \textit{Ldoor open}, and \textit{Micro open}.\\

\input{figures/benchmarks}

\section{Per-task Results of Vision-based Robot Policy Learning in Simulated Environments}
\label{sec:detailed_quan_sim}

We report the per-task results on DMC~\citep{dmc}, MetaWorld~\citep{metaworld}, and Franka Kitchen~\citep{franka_kitchen} in Table~\ref{tab:dmc}, \ref{tab:metaworld}, and \ref{tab:franka}, respectively, which decompose the aggregate results presented in the main paper.
Across all three benchmarks and all four visual representations (CLIP~\citep{clip}, SigLIP~\citep{siglip}, DINOv2~\citep{oquab2023dinov2}, and Theia~\citep{theia}), SPARK improves the reported average performance over the corresponding baselines.
Moreover, the negative per-task changes are isolated to a small number of backbone-task pairs and are outweighed by improvements elsewhere.
This suggests that the aggregate improvements are not driven by a single outlier task, but instead reflect a consistent benefit across tasks and representations.

\input{tables/quan_comp_dmc}
\input{tables/quan_comp_metaworld}

\input{tables/quan_comp_franka}

\begin{table}[h]
\caption{\textbf{Comparison with representation adaptation methods.} All methods are initialized from SigLIP~\citep{siglip}. The success rates (\%) are reported for all the tasks on Franka Kitchen~\citep{franka_kitchen}.}
\label{tab:appx_adapt}
\centering
\small
\tabcolsep=.5em
\begin{tabular}{l|ccccc|c}
\toprule
Method & Knob1 on & Light on & Sdoor open & Ldoor open & Micro open & Average \\
\midrule
SigLIP & 17.0{\scriptsize $\pm$1.2} & 38.5{\scriptsize $\pm$1.9} & 75.5{\scriptsize $\pm$3.4} & 8.5{\scriptsize $\pm$1.0} & 16.5{\scriptsize $\pm$1.9} & 31.2{\scriptsize $\pm$0.6} \\
STP & 14.0{\scriptsize $\pm$3.7} & 34.0{\scriptsize $\pm$4.3} & 56.5{\scriptsize $\pm$6.0} & 14.5{\scriptsize $\pm$2.5} & 12.0{\scriptsize $\pm$1.6} & 26.2{\scriptsize $\pm$2.1} \\
Lossless Adaptation & 17.5{\scriptsize $\pm$5.3} & 37.5{\scriptsize $\pm$5.3} & 78.5{\scriptsize $\pm$6.6} & 9.0{\scriptsize $\pm$2.0} & 12.5{\scriptsize $\pm$3.4} & 31.0{\scriptsize $\pm$1.6} \\
\midrule
SPARK & \textbf{27.5}{\scriptsize $\pm$5.0} & \textbf{43.0}{\scriptsize $\pm$5.0} & \textbf{87.5}{\scriptsize $\pm$1.9} & \textbf{20.0}{\scriptsize $\pm$2.8} & \textbf{33.0}{\scriptsize $\pm$5.3} & \textbf{42.2}{\scriptsize $\pm$2.4} \\
\bottomrule
\end{tabular}
\end{table}

\section{Comparison to Representation Adaptation Methods}
\label{sec:adaptation}
We compare SPARK against representation adaptation methods, STP~\citep{stp} and Lossless Adaptation~\citep{sharma2023lossless}, using SigLIP~\citep{siglip} as the initial model on Franka Kitchen in Table~\ref{tab:appx_adapt}.
STP falls below the baseline on average, consistent with our view that re-training how to encode can distort the pretrained representations, while Lossless Adaptation performs on par with it.
In contrast, SPARK improves every task and outperforms both methods by a large margin, as it learns to integrate the prior knowledge into a compact state rather than re-training them.

\section{Implementation details}
\label{sec:imple}

\paragraph{Initialization and post-training setup.}
For post-training, we initialize the SPARK encoder from public foundation models, including CLIP~\citep{clip}, DINOv2~\citep{oquab2023dinov2}, SigLIP~\citep{siglip}, and Theia~\citep{theia}. Unless otherwise stated, post-training is conducted on Kinetics-400~\citep{kay2017kinetics} for 50 epochs. Following repeated sampling~\citep{feichtenhofer2021large, hoffer2020augment}, this corresponds to an effective training length of 200 epochs. We use AdamW~\citep{loshchilov2018decoupled} with a batch size of 1536 and train on dynamic scenes at a resolution of $224 \times 224$. Frames are sampled at 30 FPS, and the temporal offset between the reference and target scenes is uniformly sampled between 4 and 96 frames. We apply random resized cropping and horizontal flipping, using the same crop for both reference and target scenes. The target scene is masked with a ratio of 0.9. The decoder consists of eight Vision Transformer blocks. All remaining hyperparameters follow the default pretraining settings of the corresponding backbone.

\paragraph{Policy learning setup.}
For downstream control, each agent consists of a frozen visual encoder and a policy network trained with behavior cloning. The policy takes as input the compact visual feature extracted from the encoder, together with proprioceptive input when available. For Franka Kitchen, the policy network is implemented as a two-layer MLP with batch normalization. For MetaWorld and DMC, we follow the policy-learning protocols of \citep{cortexbench}. Unless otherwise stated, the visual encoder is frozen after post-training and only the policy network is optimized.

\paragraph{Benchmark-specific settings.}
For Franka Kitchen, we evaluate five imitation learning tasks following prior protocols~\citep{nair2022r3m, parisi22}. We use either the left or right camera at a resolution of $224 \times 224$ without depth input. Training runs for 20{,}000 steps, with online evaluation every 1{,}000 steps. We report the best success rate during training, averaged over four random seeds.
For MetaWorld and DMC, we follow the setup of \citep{cortexbench}. Proprioceptive input is provided for all benchmarks except DMC. Each agent is trained for 100 epochs and evaluated online every 5 epochs. We report normalized scores for DMC and best success rates for MetaWorld, averaged across five random seeds.

%% file: figures/benchmarks.tex
\begin{figure}[h]
    \centering
    \begin{subfigure}[b]{0.22\linewidth}
        \centering
        \includegraphics[width=\linewidth]{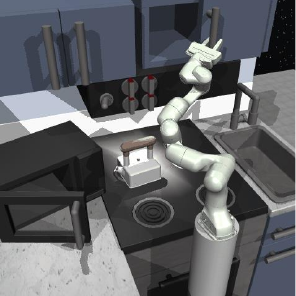}
        \caption{\footnotesize Franka Kitchen}
        \label{fig:env_franka}
    \end{subfigure}
    \hspace{0.2em}
    \begin{subfigure}[b]{0.22\linewidth}
        \centering
        \includegraphics[width=\linewidth]{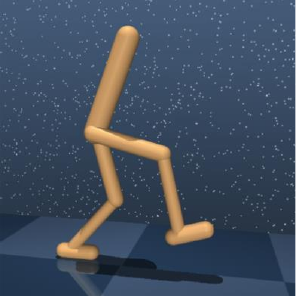}
        \caption{\footnotesize DeepMind Control}
        \label{fig:env_dmc}
    \end{subfigure}
    \hspace{0.2em}
    \begin{subfigure}[b]{0.22\linewidth}
        \centering
        \includegraphics[width=\linewidth]{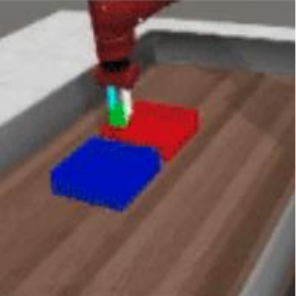}
        \caption{\footnotesize MetaWorld}
        \label{fig:env_metaworld}
    \end{subfigure}
    \caption{\textbf{Environments for evaluation.} We visualize three vision-based robot learning benchmarks employed for the validation of our method: (a) Franka Kitchen~\citep{franka_kitchen}, (b) DeepMind Control Suite~\citep{dmc}, and (c) Metaworld~\citep{metaworld}.}
    \label{fig:benchmarks}
\end{figure}

%% file: tables/quan_comp_dmc.tex
\begin{table*}[t]
\caption{\textbf{Experimental results on vision-based robot policy learning on  DeepMind Control Suite (DMC).} We report the performance of imitation learning agents on the DMC benchmark~\citep{dmc}, which are trained upon representations of CLIP~\citep{clip}, DINOv2~\citep{oquab2023dinov2}, SigLIP~\citep{siglip}, and Theia~\citep{theia} post-trained on Kinetics-400 \citep{kay2017kinetics} dataset. The normalized scores are reported for all the tasks. We report the gains of our method over the baseline.}
\label{tab:dmc}
\centering
\footnotesize
\tabcolsep=.5em
\begin{tabular}{l|l|ccccc|c}
\toprule
\multicolumn{2}{c|}{Method} & Walker-stand  & Walker-walk & Reacher-easy & Cheetah-run & Finger-spin & Average \\
\midrule
\multirow{3}{*}{CLIP}  & Baseline   & 58.1{\scriptsize $\pm$5.8} & 24.9{\scriptsize $\pm$2.8} & 75.2{\scriptsize $\pm$4.0} & 12.7{\scriptsize $\pm$2.0} & 68.4{\scriptsize $\pm$3.1} & 47.9{\scriptsize $\pm$2.1}  \\
& SPARK  & 62.0{\scriptsize $\pm$3.5} & 29.5{\scriptsize $\pm$4.5} & 75.6{\scriptsize $\pm$4.7} & 22.0{\scriptsize $\pm$2.1} & 63.7{\scriptsize $\pm$1.6} & 50.6{\scriptsize $\pm$0.9} \\
& Gain & \textcolor{darkergreen}{+ 3.9} & \textcolor{darkergreen}{+ 4.6} & \textcolor{darkergreen}{+ 0.4} & \textcolor{darkergreen}{+ 9.3} & - 4.7 & \textcolor{darkergreen}{+ 2.7}\\
\midrule
\multirow{3}{*}{SigLIP} & Baseline   & 43.9{\scriptsize $\pm$5.9} & 19.9{\scriptsize $\pm$2.8} & 74.6{\scriptsize $\pm$2.5}  & 9.0{\scriptsize $\pm$2.5} & 63.6{\scriptsize $\pm$0.4} & 42.1{\scriptsize $\pm$1.5} \\
& SPARK &  45.2{\scriptsize $\pm$6.2} & 17.5{\scriptsize $\pm$2.4} & 75.2{\scriptsize $\pm$3.0}  & 8.4{\scriptsize $\pm$2.1} & 66.8{\scriptsize $\pm$1.2} & 42.6{\scriptsize $\pm$1.8} \\
& Gain & \textcolor{darkergreen}{+ 1.3} & -2.4 & \textcolor{darkergreen}{+ 0.6} & -0.6 & \textcolor{darkergreen}{+ 3.2} & \textcolor{darkergreen}{+ 0.5}\\
\midrule
\multirow{3}{*}{DINOv2} & Baseline   & 74.2{\scriptsize $\pm$5.6} & 41.5{\scriptsize $\pm$3.4} & 76.8{\scriptsize $\pm$5.9} & 15.9{\scriptsize $\pm$3.7} & 68.3{\scriptsize $\pm$1.6} & 55.3{\scriptsize $\pm$0.4} \\
& SPARK & 78.8{\scriptsize $\pm$3.5} & 45.2{\scriptsize $\pm$5.0} & 74.3{\scriptsize $\pm$6.5} & 22.3{\scriptsize $\pm$2.9} & 70.8{\scriptsize $\pm$1.0} & 58.3{\scriptsize $\pm$2.4}  \\
& Gain & \textcolor{darkergreen}{+ 4.6} & \textcolor{darkergreen}{+ 3.7} & - 2.5 & \textcolor{darkergreen}{+ 6.4} & \textcolor{darkergreen}{+ 2.5}  & \textcolor{darkergreen}{+ 3.0}\\
\midrule
\multirow{3}{*}{Theia} & Baseline   & 74.4{\scriptsize $\pm$5.8} & 54.7{\scriptsize $\pm$8.5} & 79.1{\scriptsize $\pm$5.5} & 26.7{\scriptsize $\pm$4.4} & 69.1{\scriptsize $\pm$1.2} &  60.8{\scriptsize $\pm$2.8}\\
& SPARK & 87.7{\scriptsize $\pm$3.4} & 70.0{\scriptsize $\pm$8.0} & 83.0{\scriptsize $\pm$3.8} & 38.0{\scriptsize $\pm$4.5} & 67.5{\scriptsize $\pm$0.7} & 69.3{\scriptsize $\pm$1.4} \\
& Gain &  \textcolor{darkergreen}{+ 13.3} & \textcolor{darkergreen}{+ 15.3} & \textcolor{darkergreen}{+ 3.9} & \textcolor{darkergreen}{+ 11.3} & - 1.6 & \textcolor{darkergreen}{+ 8.5}\\
\bottomrule
\end{tabular}
\end{table*}

%% file: tables/quan_comp_metaworld.tex
\begin{table*}[t]
\caption{\textbf{Experimental results on vision-based robot policy learning on MetaWorld.} We report the performance of imitation learning agents on MetaWorld \citep{metaworld}, which are trained upon representations of CLIP~\citep{clip}, DINOv2~\citep{oquab2023dinov2}, SigLIP~\citep{siglip}, and Theia~\citep{theia} post-trained on Kinetics-400 \citep{kay2017kinetics} dataset. The success rates (\%) are reported for all the tasks. We report the gains of our method over the baseline.}
\label{tab:metaworld}
\centering
\footnotesize
\tabcolsep=.5em
\begin{tabular}{l|l|ccccc|c}
\toprule
\multicolumn{2}{c|}{Method} & assembly  & bin-picking & button-press & drawer-open & hammer & Average \\
\midrule
\multirow{3}{*}{CLIP}  & Baseline   & 55.2{\scriptsize $\pm$6.6} & 63.2{\scriptsize $\pm$7.7} & 58.4{\scriptsize $\pm$13.4} & 100.0{\scriptsize $\pm$0.0} & 87.2{\scriptsize $\pm$15.6} & 72.8{\scriptsize $\pm$1.1}  \\
& SPARK  & 72.0{\scriptsize $\pm$12.0} & 63.2{\scriptsize $\pm$7.0} & 58.8{\scriptsize $\pm$14.5} & 100.0{\scriptsize $\pm$0.0} & 94.4{\scriptsize $\pm$6.1} & 77.7{\scriptsize $\pm$7.0} \\
& Gain & \textcolor{darkergreen}{+ 16.8} & + 0.0 & \textcolor{darkergreen}{+ 0.4} & 0.0 & \textcolor{darkergreen}{+ 7.2} & \textcolor{darkergreen}{+ 4.9}\\
\midrule
\multirow{3}{*}{SigLIP} & Baseline   & 69.6{\scriptsize $\pm$9.2} & 52.0{\scriptsize $\pm$12.3} & 59.2{\scriptsize $\pm$15.6} & 100.0{\scriptsize $\pm$0.0} & 84.0{\scriptsize $\pm$9.4} & 73.0{\scriptsize $\pm$7.6} \\
& SPARK & 74.6{\scriptsize $\pm$10.3} & 60.1{\scriptsize $\pm$9.5} & 58.5{\scriptsize $\pm$12.7} & 100.0{\scriptsize $\pm$0.0} & 91.5{\scriptsize $\pm$5.8} & 76.9{\scriptsize $\pm$6.8}  \\
& Gain & \textcolor{darkergreen}{+ 5.0} & \textcolor{darkergreen}{+ 8.1} & -0.7 & 0.0 &\textcolor{darkergreen}{+ 7.5} & \textcolor{darkergreen}{+ 3.9}\\
\midrule
\multirow{3}{*}{DINOv2} & Baseline   & 64.8{\scriptsize $\pm$5.9} & 70.4{\scriptsize $\pm$14.9} & 64.0{\scriptsize $\pm$19.4} & 100.0{\scriptsize $\pm$0.0} & 92.8{\scriptsize $\pm$5.9} & 78.4 {\scriptsize $\pm$7.7} \\
& SPARK & 80.8{\scriptsize $\pm$5.2} & 67.2{\scriptsize $\pm$7.2} & 68.8{\scriptsize $\pm$16.3} & 100.0{\scriptsize $\pm$0.0} & 99.2{\scriptsize $\pm$1.8} & 83.2 {\scriptsize $\pm$5.5} \\
& Gain & \textcolor{darkergreen}{+ 16.0} & - 3.2 & \textcolor{darkergreen}{+ 4.8} & 0.0 & \textcolor{darkergreen}{+ 6.4}  & \textcolor{darkergreen}{+ 4.8} \\
\midrule
\multirow{3}{*}{Theia} & Baseline   & 88.8{\scriptsize $\pm$8.7} & 80.8{\scriptsize $\pm$4.4} & 79.2{\scriptsize $\pm$5.2} & 100.0{\scriptsize $\pm$0.0} & 96.0{\scriptsize $\pm$5.7} & 89.0{\scriptsize $\pm$2.8}\\
& SPARK & 87.2{\scriptsize $\pm$5.9} & 81.6{\scriptsize $\pm$6.1} & 86.4{\scriptsize $\pm$9.6} & 100.0{\scriptsize $\pm$0.0} & 100.0{\scriptsize $\pm$0.0}  & 89.2{\scriptsize $\pm$7.4} \\
& Gain &  -1.6 & \textcolor{darkergreen}{+ 0.8} & \textcolor{darkergreen}{+ 7.2} & 0.0 & \textcolor{darkergreen}{+ 4.0} & \textcolor{darkergreen}{+ 0.2}    \\
\bottomrule
\end{tabular}
\end{table*}

%% file: tables/quan_comp_franka.tex
\begin{table*}[t]
\caption{\textbf{Experimental results on vision-based robot policy learning on Franka Kitchen.} We report the performance of imitation learning agents on Franka Kitchen \citep{franka_kitchen}, which are trained upon representations of CLIP~\citep{clip}, DINOv2~\citep{oquab2023dinov2}, SigLIP~\citep{siglip}, and Theia~\citep{theia} post-trained on Kinetics-400 \citep{kay2017kinetics} dataset. The success rates (\%) are reported for all the tasks. We report the gains of our method over the baselines.}
\label{tab:franka}
\centering
\small
\tabcolsep=.5em
\begin{tabular}{l|l|ccccc|c}
\toprule
\multicolumn{2}{c|}{Method} & Knob1 on & Light on & Sdoor open & Ldoor open & Micro open & Average \\
\midrule
\multirow{3}{*}{CLIP}  & Baseline   & 23.0{\scriptsize $\pm$2.6} & 29.5{\scriptsize $\pm$4.7} & 69.5{\scriptsize $\pm$4.4} & 13.5{\scriptsize $\pm$3.4} & 22.0{\scriptsize $\pm$2.8} & 31.5{\scriptsize $\pm$2.4}  \\
& SPARK  & 31.5{\scriptsize $\pm$1.9} & 38.5{\scriptsize $\pm$6.0} & 74.5{\scriptsize $\pm$11.4} & 18.0{\scriptsize $\pm$2.0} & 29.0{\scriptsize $\pm$6.2} & 39.8{\scriptsize $\pm$5.6} \\
& Gain & \textcolor{darkergreen}{+ 8.5} & \textcolor{darkergreen}{+ 9.0} & \textcolor{darkergreen}{+ 5.0} & \textcolor{darkergreen}{+ 4.5} & \textcolor{darkergreen}{+ 7.0} & \textcolor{darkergreen}{+ 8.3}\\
\midrule
\multirow{3}{*}{SigLIP} & Baseline   & 17.0{\scriptsize $\pm$1.2} & 38.5{\scriptsize $\pm$1.9} & 75.5{\scriptsize $\pm$3.4} & 8.5{\scriptsize $\pm$1.0} & 16.5{\scriptsize $\pm$1.9} & 31.2{\scriptsize $\pm$0.6} \\
& SPARK & 27.5{\scriptsize $\pm$5.0} & 43.0{\scriptsize $\pm$5.0} & 87.5{\scriptsize $\pm$1.9} & 20.0{\scriptsize $\pm$2.8} & 33.0{\scriptsize $\pm$5.3} & 42.2{\scriptsize $\pm$2.4}  \\
& Gain & \textcolor{darkergreen}{+ 10.5} & \textcolor{darkergreen}{+ 4.5} & \textcolor{darkergreen}{+ 12.0} & \textcolor{darkergreen}{+ 11.5} & \textcolor{darkergreen}{+ 16.5} & \textcolor{darkergreen}{+ 11.0}\\
\midrule
\multirow{3}{*}{DINOv2} & Baseline   & 26.0{\scriptsize $\pm$4.3} & 38.5{\scriptsize $\pm$2.5} & 71.0{\scriptsize $\pm$6.8} & 9.5{\scriptsize $\pm$2.5} & 37.0{\scriptsize $\pm$6.2} & 36.4 {\scriptsize $\pm$2.7} \\
& SPARK & 21.0{\scriptsize $\pm$4.2} & 51.0{\scriptsize $\pm$8.4} & 81.5{\scriptsize $\pm$2.5} & 17.0{\scriptsize $\pm$2.0} & 33.0{\scriptsize $\pm$3.5} & 40.7{\scriptsize $\pm$2.7}  \\
& Gain & - 5.0 & \textcolor{darkergreen}{+ 12.5} & \textcolor{darkergreen}{+ 10.5} & \textcolor{darkergreen}{+ 7.5} & - 4.0  & \textcolor{darkergreen}{+ 4.3}\\
\midrule
\multirow{3}{*}{Theia} & Baseline & 33.5{\scriptsize $\pm$10.5} & 35.5{\scriptsize $\pm$6.6} & 82.5{\scriptsize $\pm$4.7} & 27.5{\scriptsize $\pm$4.7} & 14.0{\scriptsize $\pm$2.0} & 40.1{\scriptsize $\pm$2.3} \\
& SPARK & 35.0{\scriptsize $\pm$2.6} & 56.0{\scriptsize $\pm$6.5} & 75.0{\scriptsize $\pm$5.3} & 34.5{\scriptsize $\pm$2.5} & 21.0{\scriptsize $\pm$2.6} & 44.3{\scriptsize $\pm$2.6} \\
& Gain & \textcolor{darkergreen}{+ 1.5} & \textcolor{darkergreen}{+ 20.5} & - 7.5 & \textcolor{darkergreen}{+ 7.0} & \textcolor{darkergreen}{+ 7.0} & \textcolor{darkergreen}{+ 4.2} \\
\bottomrule
\end{tabular}
\end{table*}

%% file: main.bib
@String(IJCV = {Int. J. Comput. Vis.})

@String(CVPR= {IEEE Conf. Comput. Vis. Pattern Recog.})

@String(ICCV= {Int. Conf. Comput. Vis.})

@String(ICLR = {Int. Conf. Learn. Represent.})

@String(IJCV  = {IJCV})

@String(CVPR  = {CVPR})

@String(ICCV  = {ICCV})

@String(ICLR  = {ICLR})

@inproceedings{theia,
  title     = {{Theia}: Distilling Diverse Vision Foundation Models for Robot Learning},
  author    = {Shang, Jinghuan and Schmeckpeper, Karl and May, Brandon B. and Minniti, Maria Vittoria and Kelestemur, Tarik and Watkins, David and Herlant, Laura},
  booktitle = {Conference on Robot Learning (CoRL)},
  year      = {2024}
}

@inproceedings{simclr,
  title={A simple framework for contrastive learning of visual representations},
  author={Chen, Ting and Kornblith, Simon and Norouzi, Mohammad and Hinton, Geoffrey},
  booktitle={International Conference on Machine Learning},
  year={2020},
}

@InProceedings{moco,
author = {He, Kaiming and Fan, Haoqi and Wu, Yuxin and Xie, Saining and Girshick, Ross},
title = {Momentum Contrast for Unsupervised Visual Representation Learning},
booktitle = {Proceedings of the IEEE/CVF Conference on Computer Vision and Pattern Recognition (CVPR)},
month = {June},
year = {2020}
}

@Article{mocov2,
  author  = {Xinlei Chen and Haoqi Fan and Ross Girshick and Kaiming He},
  title   = {Improved Baselines with Momentum Contrastive Learning},
  journal = {arXiv preprint arXiv:2003.04297},
  year    = {2020},
}

@InProceedings{mocov3,
    author    = {Chen, Xinlei and Xie, Saining and He, Kaiming},
    title     = {An Empirical Study of Training Self-Supervised Vision Transformers},
    booktitle = {Proceedings of the IEEE/CVF International Conference on Computer Vision (ICCV)},
    month     = {October},
    year      = {2021},
    pages     = {9640-9649}
}

@inproceedings{dino,
  title={Emerging Properties in Self-Supervised Vision Transformers},
  author={Caron, Mathilde and Touvron, Hugo and Misra, Ishan and J{\'e}gou, Herv{\'e} and Mairal, Julien and Bojanowski, Piotr and Joulin, Armand},
  booktitle={Proceedings of the IEEE/CVF International Conference on Computer Vision},
  year={2021}
}

@inproceedings{mae,
  title={Masked Autoencoders Are Scalable Vision Learners},
  author={He, Kaiming and Chen, Xinlei and Xie, Saining and Li, Yanghao and Doll{\'a}r, Piotr and Girshick, Ross},
  booktitle={Proceedings of the IEEE/CVF Conference on Computer Vision and Pattern Recognition},
  year={2022}
}

@inproceedings{siammae,
  title={Siamese Masked Autoencoders},
  author={Gupta, Agrim and Wu, Jiajun and Deng, Jia and Fei-Fei, Li},
  booktitle={Advances in Neural Information Processing Systems},
  year={2023}
}

@article{kay2017kinetics,
  title={The {K}inetics human action video dataset},
  author={Kay, Will and Carreira, Joao and Simonyan, Karen and Zhang, Brian and Hillier, Chloe and Vijayanarasimhan, Sudheendra and Viola, Fabio and Green, Tim and Back, Trevor and Natsev, Paul and others},
  journal={arXiv preprint arXiv:1705.06950},
  year={2017}
}

@article{rlbench,
  title={{RLBench}: The robot learning benchmark \& learning environment},
  author={James, Stephen and Ma, Zicong and Arrojo, David Rovick and Davison, Andrew J},
  journal={IEEE Robotics and Automation Letters},
  volume={5},
  number={2},
  pages={3019--3026},
  year={2020},
  publisher={IEEE}
}

@InProceedings{franka_kitchen,
  title = 	 {Relay Policy Learning: Solving Long-Horizon Tasks via Imitation and Reinforcement Learning},
  author =       {Gupta, Abhishek and Kumar, Vikash and Lynch, Corey and Levine, Sergey and Hausman, Karol},
  booktitle = 	 {Proceedings of the Conference on Robot Learning},
  pages = 	 {1025--1037},
  year = 	 {2020},
  volume = 	 {100},
  series = 	 {Proceedings of Machine Learning Research},
  month = 	 {30 Oct--01 Nov},
  publisher =    {PMLR},
  url = 	 {https://proceedings.mlr.press/v100/gupta20a.html},
}

@inproceedings{
cortexbench,
title={Where are we in the search for an Artificial Visual Cortex for Embodied Intelligence?},
author={Arjun Majumdar and Karmesh Yadav and Sergio Arnaud and Yecheng Jason Ma and Claire Chen and Sneha Silwal and Aryan Jain and Vincent-Pierre Berges and Tingfan Wu and Jay Vakil and Pieter Abbeel and Jitendra Malik and Dhruv Batra and Yixin Lin and Oleksandr Maksymets and Aravind Rajeswaran and Franziska Meier},
booktitle = {Advances in Neural Information Processing Systems},
volume = {36},
year={2023},
url={https://openreview.net/forum?id=6qLzQeFGio}
}

@article{dmc,
         title = {{dm\_control}: Software and tasks for continuous control},
         journal = {Software Impacts},
         volume = {6},
         pages = {100022},
         year = {2020},
         doi = {10.1016/j.simpa.2020.100022},
         author = {Saran Tunyasuvunakool and Alistair Muldal and Yotam Doron and
                   Siqi Liu and Steven Bohez and Josh Merel and Tom Erez and
                   Timothy Lillicrap and Nicolas Heess and Yuval Tassa},
}

@inproceedings{metaworld,
  title={{Meta-World}: A benchmark and evaluation for multi-task and meta reinforcement learning},
  author={Yu, Tianhe and Quillen, Deirdre and He, Zhanpeng and Julian, Ryan and Hausman, Karol and Finn, Chelsea and Levine, Sergey},
  booktitle={Conference on Robot Learning},
  year={2020},
}

@InProceedings{parisi22,
  title = 	 {The Unsurprising Effectiveness of Pre-Trained Vision Models for Control},
  author =       {Parisi, Simone and Rajeswaran, Aravind and Purushwalkam, Senthil and Gupta, Abhinav},
  booktitle = 	 {Proceedings of the 39th International Conference on Machine Learning},
  pages = 	 {17359--17371},
  year = 	 {2022},
  volume = 	 {162},
  series = 	 {Proceedings of Machine Learning Research},
  month = 	 {17--23 Jul},
  publisher =    {PMLR},
  url = 	 {https://proceedings.mlr.press/v162/parisi22a.html},
}

@inproceedings{hoffer2020augment,
  title={Augment your batch: Improving generalization through instance repetition},
  author={Hoffer, Elad and Ben-Nun, Tal and Hubara, Itay and Giladi, Niv and Hoefler, Torsten and Soudry, Daniel},
  booktitle={Proceedings of the IEEE/CVF Conference on Computer Vision and Pattern Recognition},
  year={2020}
}

@inproceedings{feichtenhofer2021large,
  title={A large-scale study on unsupervised spatiotemporal representation learning},
  author={Feichtenhofer, Christoph and Fan, Haoqi and Xiong, Bo and Girshick, Ross and He, Kaiming},
  booktitle={Proceedings of the IEEE/CVF Conference on Computer Vision and Pattern Recognition},
  pages={3299--3309},
  year={2021}
}

@inproceedings{
    loshchilov2018decoupled,
    title={Decoupled Weight Decay Regularization},
    author={Ilya Loshchilov and Frank Hutter},
    booktitle={International Conference on Learning Representations},
    year={2019},
}

@inproceedings{zeng2024mpi,
  title     = {Learning Manipulation by Predicting Interaction},
  author    = {Zeng, Jia and Bu, Qingwen and Wang, Bangjun and Xia, Wenke and Chen, Li and Dong, Hao and Song, Haoming and Wang, Dong and Hu, Di and Luo, Ping and Cui, Heming and Zhao, Bin and Li, Xuelong and Qiao, Yu and Li, Hongyang},
  booktitle = {Proceedings of Robotics: Science and Systems (RSS)},
  year      = {2024}
}

@inproceedings{karamcheti2023voltron,
  title={Language-Driven Representation Learning for Robotics},
  author={Siddharth Karamcheti and Suraj Nair and Annie S. Chen and Thomas Kollar and Chelsea Finn and Dorsa Sadigh and Percy Liang},
  booktitle={Robotics: Science and Systems (RSS)},
  year={2023}
}

@inproceedings{nair2022r3m,
title={{R3M}: A Universal Visual Representation for Robot Manipulation},
author={Suraj Nair and Aravind Rajeswaran and Vikash Kumar and Chelsea Finn and Abhinav Gupta},
booktitle={6th Annual Conference on Robot Learning},
year={2022},
url={https://openreview.net/forum?id=tGbpgz6yOrI}
}

@inproceedings{data4robotics,
      title={An Unbiased Look at Datasets for Visuo-Motor Pre-Training},
      author={Dasari, Sudeep and Srirama, Mohan Kumar and Jain, Unnat and Gupta, Abhinav},
      booktitle={Conference on Robot Learning},
      year={2023},
      organization={PMLR}
    }

@InProceedings{radosavovic2023mvp,
  title = 	 {Real-World Robot Learning with Masked Visual Pre-training},
  author =       {Radosavovic, Ilija and Xiao, Tete and James, Stephen and Abbeel, Pieter and Malik, Jitendra and Darrell, Trevor},
  booktitle = 	 {Proceedings of The 6th Conference on Robot Learning},
  pages = 	 {416--426},
  year = 	 {2023},
  volume = 	 {205},
  series = 	 {Proceedings of Machine Learning Research},
  month = 	 {14--18 Dec},
  publisher =    {PMLR},
  url = 	 {https://proceedings.mlr.press/v205/radosavovic23a.html},
}

@inproceedings{clip,
  title={Learning transferable visual models from natural language supervision},
  author={Radford, Alec and Kim, Jong Wook and Hallacy, Chris and Ramesh, Aditya and Goh, Gabriel and Agarwal, Sandhini and Sastry, Girish and Askell, Amanda and Mishkin, Pamela and Clark, Jack and others},
  booktitle={International Conference on Machine Learning},
  pages={8748--8763},
  year={2021},
  organization={PMLR}
}

@article{
oquab2023dinov2,
title={{DINO}v2: Learning Robust Visual Features without Supervision},
author={Maxime Oquab and Timoth{\'e}e Darcet and Th{\'e}o Moutakanni and Huy V. Vo and Marc Szafraniec and Vasil Khalidov and Pierre Fernandez and Daniel Haziza and Francisco Massa and Alaaeldin El-Nouby and Mido Assran and Nicolas Ballas and Wojciech Galuba and Russell Howes and Po-Yao Huang and Shang-Wen Li and Ishan Misra and Michael Rabbat and Vasu Sharma and Gabriel Synnaeve and Hu Xu and Herve J{\'e}gou and Julien Mairal and Patrick Labatut and Armand Joulin and Piotr Bojanowski},
journal={Transactions on Machine Learning Research},
year={2024},
url={https://openreview.net/forum?id=a68SUt6zFt},
}

@InProceedings{ego4d,
    author    = {Grauman, Kristen and Westbury, Andrew and Byrne, Eugene and Chavis, Zachary and Furnari, Antonino and Girdhar, Rohit and Hamburger, Jackson and Jiang, Hao and Liu, Miao and Liu, Xingyu and Martin, Miguel and Nagarajan, Tushar and Radosavovic, Ilija and Ramakrishnan, Santhosh Kumar and Ryan, Fiona and Sharma, Jayant and Wray, Michael and Xu, Mengmeng and Xu, Eric Zhongcong and Zhao, Chen and Bansal, Siddhant and Batra, Dhruv and Cartillier, Vincent and Crane, Sean and Do, Tien and Doulaty, Morrie and Erapalli, Akshay and Feichtenhofer, Christoph and Fragomeni, Adriano and Fu, Qichen and Gebreselasie, Abrham and Gonz\'alez, Cristina and Hillis, James and Huang, Xuhua and Huang, Yifei and Jia, Wenqi and Khoo, Weslie and Kol\'a\v{r}, J\'achym and Kottur, Satwik and Kumar, Anurag and Landini, Federico and Li, Chao and Li, Yanghao and Li, Zhenqiang and Mangalam, Karttikeya and Modhugu, Raghava and Munro, Jonathan and Murrell, Tullie and Nishiyasu, Takumi and Price, Will and Ruiz, Paola and Ramazanova, Merey and Sari, Leda and Somasundaram, Kiran and Southerland, Audrey and Sugano, Yusuke and Tao, Ruijie and Vo, Minh and Wang, Yuchen and Wu, Xindi and Yagi, Takuma and Zhao, Ziwei and Zhu, Yunyi and Arbel\'aez, Pablo and Crandall, David and Damen, Dima and Farinella, Giovanni Maria and Fuegen, Christian and Ghanem, Bernard and Ithapu, Vamsi Krishna and Jawahar, C. V. and Joo, Hanbyul and Kitani, Kris and Li, Haizhou and Newcombe, Richard and Oliva, Aude and Park, Hyun Soo and Rehg, James M. and Sato, Yoichi and Shi, Jianbo and Shou, Mike Zheng and Torralba, Antonio and Torresani, Lorenzo and Yan, Mingfei and Malik, Jitendra},
    title     = {{Ego4D}: Around the World in 3,000 Hours of Egocentric Video},
    booktitle = {Proceedings of the IEEE/CVF Conference on Computer Vision and Pattern Recognition (CVPR)},
    month     = {June},
    year      = {2022},
    pages     = {18995-19012}
}

@InProceedings{ssv2,
author = {Goyal, Raghav and Ebrahimi Kahou, Samira and Michalski, Vincent and Materzynska, Joanna and Westphal, Susanne and Kim, Heuna and Haenel, Valentin and Fruend, Ingo and Yianilos, Peter and Mueller-Freitag, Moritz and Hoppe, Florian and Thurau, Christian and Bax, Ingo and Memisevic, Roland},
title = {The "Something Something" Video Database for Learning and Evaluating Visual Common Sense},
booktitle = {Proceedings of the IEEE International Conference on Computer Vision (ICCV)},
month = {Oct},
year = {2017}
}

@InProceedings{siglip,
    author    = {Zhai, Xiaohua and Mustafa, Basil and Kolesnikov, Alexander and Beyer, Lucas},
    title     = {Sigmoid Loss for Language Image Pre-Training},
    booktitle = {Proceedings of the IEEE/CVF International Conference on Computer Vision (ICCV)},
    month     = {October},
    year      = {2023},
    pages     = {11975-11986}
}

@article{siglip2,
  title={{SigLIP} 2: Multilingual Vision-Language Encoders with Improved Semantic Understanding, Localization, and Dense Features},
  author={Tschannen, Michael and Gritsenko, Alexey and Wang, Xiao and Naeem, Muhammad Ferjad and Alabdulmohsin, Ibrahim and Parthasarathy, Nikhil and Evans, Talfan and Beyer, Lucas and Xia, Ye and Mustafa, Basil and H\'enaff, Olivier and Harmsen, Jeremiah and Steiner, Andreas and Zhai, Xiaohua},
  year={2025},
  journal={arXiv preprint arXiv:2502.14786}
}

@inproceedings{zhou2022cocoop,
    title={Conditional Prompt Learning for Vision-Language Models},
    author={Zhou, Kaiyang and Yang, Jingkang and Loy, Chen Change and Liu, Ziwei},
    booktitle={IEEE/CVF Conference on Computer Vision and Pattern Recognition (CVPR)},
    year={2022}
}

@article{zhou2022coop,
    title={Learning to Prompt for Vision-Language Models},
    author={Zhou, Kaiyang and Yang, Jingkang and Loy, Chen Change and Liu, Ziwei},
    journal={International Journal of Computer Vision (IJCV)},
    year={2022}
}

@article{gao2021clip,
author = {Gao, Peng and Geng, Shijie and Zhang, Renrui and Ma, Teli and Fang, Rongyao and Zhang, Yongfeng and Li, Hongsheng and Qiao, Yu},
title = {{CLIP-Adapter}: Better Vision-Language Models with Feature Adapters},
year = {2024},
issue_date = {Feb 2024},
publisher = {Kluwer Academic Publishers},
journal   = {International Journal of Computer Vision},
address = {USA},
volume = {132},
number = {2},
pages = {581–595},
numpages = {15}
}

@inproceedings{
hu2022lora,
title={Lo{RA}: Low-Rank Adaptation of Large Language Models},
author={Edward J Hu and Yelong Shen and Phillip Wallis and Zeyuan Allen-Zhu and Yuanzhi Li and Shean Wang and Lu Wang and Weizhu Chen},
booktitle={International Conference on Learning Representations},
year={2022},
url={https://openreview.net/forum?id=nZeVKeeFYf9}
}

@inproceedings{
ma2022vip,
title={{VIP}: Towards Universal Visual Reward and Representation via Value-Implicit Pre-Training},
author={Yecheng Jason Ma and Shagun Sodhani and Dinesh Jayaraman and Osbert Bastani and Vikash Kumar and Amy Zhang},
booktitle={The Eleventh International Conference on Learning Representations },
year={2023},
url={https://openreview.net/forum?id=YJ7o2wetJ2}
}

@article{allal2025smollm2,
  title   = {{SmolLM2}: When Smol Goes Big -- Data-Centric Training of a Small Language Model},
  author  = {Ben Allal, Loubna and Lozhkov, Anton and Bakouch, Elie and Martín Blázquez, Gabriel and Penedo, Guilherme and Tunstall, Lewis and Marafioti, Andrés and Kydlíček, Hynek and Piqueres Lajarín, Agustín and Srivastav, Vaibhav and Lochner, Joshua and Fahlgren, Caleb and Nguyen, Xuan-Son and Fourrier, Clémentine and Burtenshaw, Ben and Larcher, Hugo and Zhao, Haojun and Zakka, Cyril and Morlon, Mathieu and Raffel, Colin and von Werra, Leandro and Wolf, Thomas},
  journal = {arXiv preprint arXiv:2502.02737},
  year    = {2025}
}

@inproceedings{liu2023libero,
 author = {Liu, Bo and Zhu, Yifeng and Gao, Chongkai and Feng, Yihao and Liu, Qiang and Zhu, Yuke and Stone, Peter},
 booktitle = {Advances in Neural Information Processing Systems},
 doi = {10.52202/075280-1939},
 pages = {44776--44791},
 publisher = {Curran Associates, Inc.},
 title = {{LIBERO}: Benchmarking Knowledge Transfer for Lifelong Robot Learning},
 volume = {36},
 year = {2023}
}

@inproceedings{liu2023llava,
  title     = {Visual Instruction Tuning},
  author    = {Liu, Haotian and Li, Chunyuan and Wu, Qingyang and Lee, Yong Jae},
  booktitle = {Advances in Neural Information Processing Systems},
  year      = {2023}
}

@article{young2014image,
  title={From image descriptions to visual denotations: New similarity metrics for semantic inference over event descriptions},
  author={Young, Peter and Lai, Alice and Hodosh, Micah and Hockenmaier, Julia},
  journal={Transactions of the Association for Computational Linguistics},
  volume={2},
  pages={67--78},
  year={2014},
  publisher={MIT Press}
}

@inproceedings{
kim2024openvla,
title={Open{VLA}: An Open-Source Vision-Language-Action Model},
author={Moo Jin Kim and Karl Pertsch and Siddharth Karamcheti and Ted Xiao and Ashwin Balakrishna and Suraj Nair and Rafael Rafailov and Ethan P Foster and Pannag R Sanketi and Quan Vuong and Thomas Kollar and Benjamin Burchfiel and Russ Tedrake and Dorsa Sadigh and Sergey Levine and Percy Liang and Chelsea Finn},
booktitle={Proc. of the 8th Annual Conference on Robot Learning (CoRL)},
year={2024},
url={https://openreview.net/forum?id=ZMnD6QZAE6}
}

@InProceedings{intelligence2025pi,
  title = 	 {$\pi_{0.5}$: a Vision-Language-Action Model with Open-World Generalization},
  author =       {Black, Kevin and Brown, Noah and Darpinian, James and Dhabalia, Karan and Driess, Danny and Esmail, Adnan and Equi, Michael Robert and Finn, Chelsea and Fusai, Niccolo and Galliker, Manuel Y. and Ghosh, Dibya and Groom, Lachy and Hausman, Karol and ichter, brian and Jakubczak, Szymon and Jones, Tim and Ke, Liyiming and LeBlanc, Devin and Levine, Sergey and Li-Bell, Adrian and Mothukuri, Mohith and Nair, Suraj and Pertsch, Karl and Ren, Allen Z. and Shi, Lucy Xiaoyang and Smith, Laura and Springenberg, Jost Tobias and Stachowicz, Kyle and Tanner, James and Vuong, Quan and Walke, Homer and Walling, Anna and Wang, Haohuan and Yu, Lili and Zhilinsky, Ury},
  booktitle = 	 {Proceedings of The 9th Conference on Robot Learning},
  pages = 	 {17--40},
  year = 	 {2025},
  volume = 	 {305},
  series = 	 {Proceedings of Machine Learning Research},
  month = 	 {27--30 Sep},
  publisher =    {PMLR},
  url = 	 {https://proceedings.mlr.press/v305/black25a.html},
}

@article{bjorck2025groot,
  title   = {{GR00T N1}: An Open Foundation Model for Generalist Humanoid Robots},
  author  = {NVIDIA and Bjorck, Johan and Castañeda, Fernando and Cherniadev, Nikita and Da, Xingye and Ding, Runyu and Fan, Linxi and Fang, Yu and Fox, Dieter and Hu, Fengyuan and Huang, Spencer and Jang, Joel and Jiang, Zhenyu and Kautz, Jan and Kundalia, Kaushil and Lao, Lawrence and Li, Zhiqi and Lin, Zongyu and Lin, Kevin and Liu, Guilin and Llontop, Edith and Magne, Loic and Mandlekar, Ajay and Narayan, Avnish and Nasiriany, Soroush and Reed, Scott and Tan, You Liang and Wang, Guanzhi and Wang, Zu and Wang, Jing and Wang, Qi and Xiang, Jiannan and Xie, Yuqi and Xu, Yinzhen and Xu, Zhenjia and Ye, Seonghyeon and Yu, Zhiding and Zhang, Ao and Zhang, Hao and Zhao, Yizhou and Zheng, Ruijie and Zhu, Yuke},
  journal = {arXiv preprint arXiv:2503.14734},
  year    = {2025}
}

@inproceedings{
zheng2026xvla,
title={X-{VLA}: Soft-Prompted Transformer as Scalable Cross-Embodiment Vision-Language-Action Model},
author={Jinliang Zheng and Jianxiong Li and Zhihao Wang and Dongxiu Liu and Xirui Kang and Yuchun Feng and Yinan Zheng and Jiayin Zou and Yilun Chen and Jia Zeng and Tai Wang and Ya-Qin Zhang and Jingjing Liu and Xianyuan Zhan},
booktitle={Proc. of the Fourteenth International Conference on Learning Representations (ICLR)},
year={2026},
url={https://openreview.net/forum?id=kt51kZH4aG}
}

@article{stp,
  title   = {Spatiotemporal Predictive Pre-training for Robotic Motor Control},
  author  = {Yang, Jiange and Liu, Bei and Fu, Jianlong and Pan, Bocheng and Wu, Gangshan and Wang, Limin},
  journal = {International Journal of Computer Vision},
  volume  = {134},
  pages   = {354},
  year    = {2026},
  doi     = {10.1007/s11263-026-02948-3}
}

@inproceedings{
sharma2023lossless,
title={Lossless Adaptation of Pretrained Vision Models For Robotic Manipulation},
author={Mohit Sharma and Claudio Fantacci and Yuxiang Zhou and Skanda Koppula and Nicolas Heess and Jon Scholz and Yusuf Aytar},
booktitle={The Eleventh International Conference on Learning Representations },
year={2023},
url={https://openreview.net/forum?id=5IND3TXJRb-}
}

@inproceedings{chen2023automae,
  title     = {Improving Masked Autoencoders by Learning Where to Mask},
  author    = {Chen, Haijian and Zhang, Wendong and Wang, Yunbo and Yang, Xiaokang},
  booktitle = {Pattern Recognition and Computer Vision (PRCV)},
  year      = {2023},
  doi       = {10.1007/978-981-99-8543-2_31}
}
